\documentclass[runningheads]{llncs}
\usepackage{graphicx}

\usepackage[table]{xcolor}

\usepackage{tikz}
\usepackage{comment}
\usepackage{amsmath,amssymb} 
\usepackage{color}

\usepackage{epsfig}
\usepackage{url}            
\usepackage{amsfonts}       
\usepackage{multirow}
\usepackage{array}
\usepackage{graphbox}
\usepackage{booktabs} 
\usepackage{wrapfig}
\usepackage{enumitem}
\usepackage{bbding}
\usepackage{float}
\usepackage{placeins}
\usepackage[accsupp]{axessibility}  

\usepackage{color}
\definecolor{mypink3}{cmyk}{0, 0.7808, 0.4429, 0.1412}

\definecolor{Gray}{gray}{0.8}

\usepackage[all]{xy}
\usepackage{amsmath}
\usepackage{wrapfig}

\usepackage{academicons}
\definecolor{orcidlogocol}{HTML}{A6CE39}

\usepackage{orcidlink}

\begin{document}
\pagestyle{headings}
\mainmatter
\def\ECCVSubNumber{303}  

\title{DecoupleNet: Decoupled Network for Domain Adaptive Semantic Segmentation} 


\titlerunning{DecoupleNet}
%
\author{Xin Lai\inst{1}\orcidlink{0000-0002-2973-3043} \and
Zhuotao Tian\inst{1}\orcidlink{0000-0003-2698-6923} \and Xiaogang Xu\inst{1}\orcidlink{0000-0002-7928-7336} \and Yingcong Chen\inst{3,4,5}\orcidlink{0000-0002-9565-8205}\thanks{Corresponding Author} \and Shu Liu\inst{2}\orcidlink{0000-0002-2903-9270} \and Hengshuang Zhao\inst{6,7}\orcidlink{0000-0001-8277-2706} \and Liwei Wang\inst{1}\orcidlink{0000-0003-3264-1294} \and
Jiaya Jia\inst{1,2}\orcidlink{0000-0002-1246-553X}}
\authorrunning{X. Lai et al.}
%
\institute{$^{1}$CUHK \quad $^{2}$SmartMore \quad $^{3}$HKUST(GZ) \quad $^{4}$HKUST \\ $^{5}$HKUST(GZ)-SmartMore Joint Lab \quad $^{6}$HKU \quad $^{7}$MIT}
\maketitle

\begin{abstract}
Unsupervised domain adaptation in semantic segmentation has been raised to alleviate the reliance on expensive pixel-wise annotations. It leverages a labeled source domain dataset as well as unlabeled target domain images to learn a segmentation network. In this paper, we observe two main issues of the existing domain-invariant learning framework. (1) Being distracted by the feature distribution alignment, the network cannot focus on the segmentation task. (2) Fitting source domain data well would compromise the target domain performance. To address these issues, we propose DecoupleNet that alleviates source domain overfitting and enables the final model to focus more on the segmentation task. Furthermore, we put forward Self-Discrimination (SD) and introduce an auxiliary classifier to learn more discriminative target domain features with pseudo labels. Finally, we propose Online Enhanced Self-Training (OEST) to contextually enhance the quality of pseudo labels in an online manner. Experiments show our method outperforms existing state-of-the-art methods, and extensive ablation studies verify the effectiveness of each component. Code is available at \url{https://github.com/dvlab-research/DecoupleNet}.

\end{abstract}

\section{Introduction}

Semantic segmentation has made tremendous progress in recent years and it has significantly benefited plenty of applications. However, satisfying performance highly relies on pixel-wise annotations. In this work, to alleviate the data-reliance issue, we focus on unsupervised domain adaptation (UDA), aiming to learn a segmentation network with a labeled source domain dataset (usually a physically synthetic dataset) and an unlabeled target domain dataset. 

\begin{figure}[t]
\begin{center}
\includegraphics[width=1.0\linewidth]{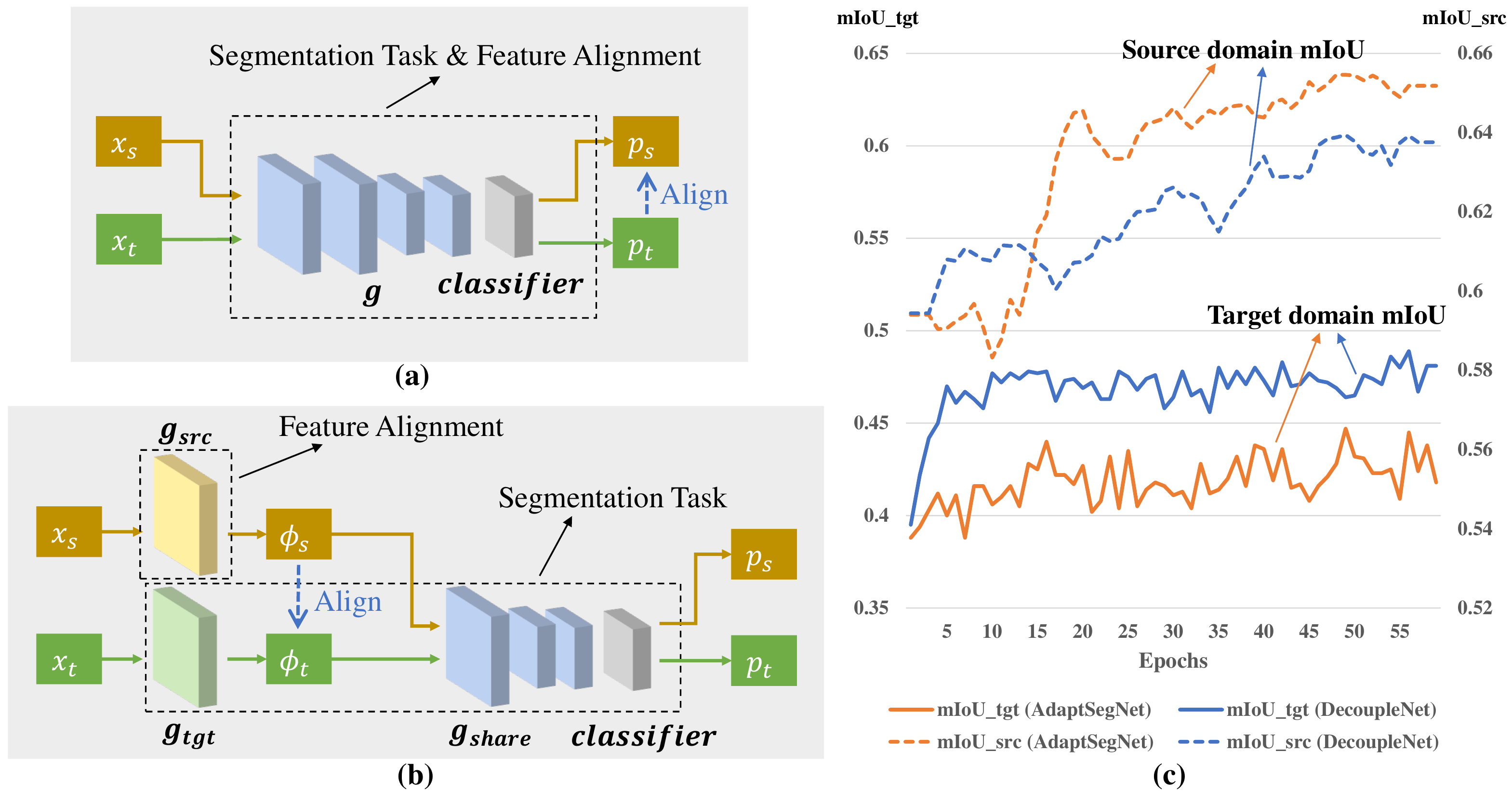}
\end{center}
\caption{(a) Domain-invariant learning. (b) Our proposed DecoupleNet. The original encoder $\boldsymbol g$ is split into $\boldsymbol g_{tgt}$ and $\boldsymbol g_{share}$. Also, $\boldsymbol g_{src}$ and $\boldsymbol g_{tgt}$ share the same architecture but not the parameters. The source-domain shallow features $\phi_s$ are aligned towards $\phi_t$ with an adversarial loss. During inference, $\boldsymbol g = \boldsymbol g_{share} \circ \boldsymbol g_{tgt}$ is used, and $\boldsymbol g_{src}$ is simply discarded. (c) Plot of the validation mIoU on source domain data (mIoU\_src) by the dashed line and on target domain data (mIoU\_tgt) by the solid line. We compare our DecoupleNet (blue line) with a representative domain-invariant learning method, i.e., AdaptSegNet~\cite{adaptnet} (orange line)}
\label{fig:intro}
\end{figure}

Due to ``domain shift"~\cite{domain_shift1,domain_shift2} between the source and target domains, directly adopting the model trained on the source domain causes much performance degradation on the target one. To minimize domain shift, domain-invariant learning~\cite{adaptnet,adapt_patch,advent,clan,ssf,fada} was put forward to align distributions of source and target features. Specifically, the features or predictions from different domains are aligned with a discriminator in the manner of adversarial learning, as shown in Fig.~\ref{fig:intro}(a). The discriminator learns to distinguish between source and target features, while the segmentation network learns to generate features that can fool the discriminator.

Domain-invariant learning alleviates domain shift. However, we still observe the following two problems.

\noindent(1) \textit{Tasks entanglement.} The feature distribution alignment and the segmentation task are conducted simultaneously in a single network, as shown in Fig.~\ref{fig:intro}(a). Being distracted by feature distribution alignment, the network cannot focus on semantic segmentation, leading to inferior performance.

\noindent(2) \textit{Source domain overfitting.} Since the training objective involves cross-entropy loss that minimizes the error on the source domain data, the trained model would fit the source domain data well, as shown in Fig.~\ref{fig:intro}(c). However, in UDA, we only care about the performance on the target domain, regardless of how it performs on the source domain. Moreover, as we will discuss in Sec.~\ref{sec:motivation}, fitting the source domain very well would contrarily compromise the target domain performance.

Based on these two observations, we design DecoupleNet to decouple feature distribution alignment and the segmentation task. As shown in Fig.~\ref{fig:intro}(b), we introduce a copy of shallow encoder layers for the source domain, i.e., $\boldsymbol g_{src}$, during training. Our goal is to let $\boldsymbol g_{src}$ conduct feature distribution alignment, such that the final model $\boldsymbol g = \boldsymbol g_{share} \circ \boldsymbol g_{tgt}$ focuses more on the downstream segmentation task. Also, it is notable that $\boldsymbol g_{src}$ is simply discarded during inference, and it only incurs negligible computational costs during training as shown in the supplementary material. 

With our new design, the issue of \textit{tasks entanglement} can be addressed, as shown in Fig.~\ref{fig:intro}(b). 
Moreover, during training, we only require the model $\boldsymbol g_{share} \circ \boldsymbol g_{src}$ to fit well on the source domain, but never ask the final model $\boldsymbol g = \boldsymbol g_{share} \circ \boldsymbol g_{tgt}$ to do so. Thus, the final model avoids overfitting in the source domain. As shown in Fig.~\ref{fig:intro}(c), compared to the domain-invariant method (AdaptSegNet~\cite{adaptnet}), DecoupleNet alleviates the \textit{source domain overfitting} problem, and boosts the target domain performance.

In addition, in order to learn more discriminative features for the target domain, we propose the Self-Discrimination (SD) technique by virtue of pseudo labels. Unlike most self-training-based methods~\cite{zou2018unsupervised,shin2020two,mei2020instance,zhang2021prototypical,yang2020label,kim2020learning,fada}, SD does not need another training phase to re-train the whole network from scratch. Instead, pseudo labels are generated at each training iteration and can be employed as an additional supervision in an online manner.
Given the fact that directly adopting the noisy pseudo labels to supervise itself could corrupt the existing classifier, we introduce an auxiliary classifier during training to prevent the contamination. 
 
Finally, we propose Online Enhanced Self-Training (OEST) to further boost the performance by extending DecoupleNet to a multi-stage training paradigm. Most existing self-training-based methods~\cite{zou2018unsupervised,shin2020two,mei2020instance,yang2020label,kim2020learning,fada} directly use the generated pseudo labels without updating them in the re-training process. Contrarily, at each training iteration, OEST updates the pseudo labels by fusing current contextually enhanced predictions, which effectively improves the quality of pseudo labels.

In summary, our contribution is threefold.
\begin{itemize}

    \item We propose DecoupleNet to decouple feature distribution alignment and semantic segmentation. This enables the network to get rid of tasks entanglement and focus more on the segmentation task.
    
    \item To learn more discriminative features, we put forward Self-Discrimination by introducing an auxiliary classifier. Moreover, we propose Online Enhanced Self-Training to contextually enhance the quality of pseudo labels.
    
    \item Experiments show that our approach outperforms existing state-of-the-art methods by a large margin.
    Also, extensive ablation studies verify the effectiveness of each component in our method.
\end{itemize}

\section{Related Work}

\subsubsection{Semantic segmentation.}
Semantic segmentation aims to assign a class label to evey pixel in an image. FCN~\cite{fcn} is a classic semantic segmentation network, which puts forward a fully-convolutional network. Considering that the final output size of FCN is smaller than the input, methods based on encoder-decoder structures~\cite{deconvnet,segnet,unet} are proposed to refine the output. Though the high-level feature has already encoded the semantic information, it cannot well capture the long-range relationship. Dilated convolution~\cite{deeplab,dilation}, global pooling~\cite{parsenet}, pyramid pooling~\cite{pspnet,icnet,denseaspp} and attention mechanism~\cite{danet,ccnet,psanet,asymmetric_nonlocal} are used to better incorporate the context. 
Despite the success, all the models need annotations to accomplish training, which costs much human effort.

\subsubsection{Unsupervised domain adaptation.}
Unsupervised dmain adaptation~\cite{gopalan2011domain} intends to alleviate the data-reliance with a labeled dataset from a different domain. 
Distance-based methods~\cite{long2015learning,long2016unsupervised,long2017deep,tzeng2014deep,sun2016deep,li2021semantic,wei2021metaalign,liu2021recursively} minimize the distribution distance such as MMD~\cite{tzeng2014deep} between the source and target domain. With the development of Generative Adversarial Network (GAN)~\cite{goodfellow2014generative}, adversarial learning methods~\cite{ganin2015unsupervised,tzeng2017adversarial,xie2018learning,hong2018conditional,liu2016coupled,xu2020adversarial,cicek2019unsupervised,deng2019cluster,liu2019transferable,volpi2018adversarial,kurmi2019attending,ma2019gcan,zhang2019domain,chen2019progressive,kang2018deep,awais2021adversarial,liu2021adversarial} get popular to align the marginal or conditional feature distributions between the source and target domains. Also, methods of \cite{cui2020gradually,peng2020domain2vec} factorize the feature into domain-specific and domain-agnostic features.

\subsubsection{UDA in Semantic Segmentation.}
\label{sec:uda_semseg}
AdaptSegNet~\cite{adaptnet} employs adversarial learning to align predictions between the source and target domain in the output space and method of \cite{luo2019significance} makes further improvement. Patch-level information is used in \cite{adapt_patch} to improve the performance and contextual relationship is considered in \cite{huang2020contextual,kang2020pixel} explicitly. \cite{zhang2019category,wang2021domain,pandey2020unsupervised} directly minimize the feature distance. In \cite{advent,yang2020adversarial,lee2019drop,truong2021bimal,yang2021exploring,liu2021domain,lai2021semi,zhang2022unsupervised,sohn2020fixmatch}, semi-supervised learning methods, such as entropy minimization, adding perturbation, contrastive learning and randomly dropout, further boost performance. \cite{fada,ssf,clan} align class-conditioned feature distribution. Methods of \cite{li2016revisiting,maria2017autodial,mancini2018boosting,wonng2019domain} provide distinct processing for features from different domains on some modules.
On the other hand, image-to-image translation methods are considered in \cite{cycada,bdl,gong2019dlow,yang2020label,cheng2021dual,yang2020fda}. Recently, self-training-based methods~\cite{zou2018unsupervised,shin2020two,mei2020instance,zhang2021prototypical,yang2020label,kim2020learning,fada,guo2021metacorrection,guizilini2021geometric} re-train the network with the pseudo labels generated from the initial network, yielding considerable improvement. 


\section{Our Method}
In this section, we first introduce the preliminary in Sec.~\ref{sec:preliminary}. Then, the key observations are presented as our motivation in Sec.~\ref{sec:motivation}. Afterwards, DecoupleNet, SD and OEST are elaborated in Sec.~\ref{sec:arch}, Sec.~\ref{sec:sd} and Sec.~\ref{sec:oest}, respectively.

\begin{figure}[t]
\begin{center}
\includegraphics[width=1.0\linewidth]{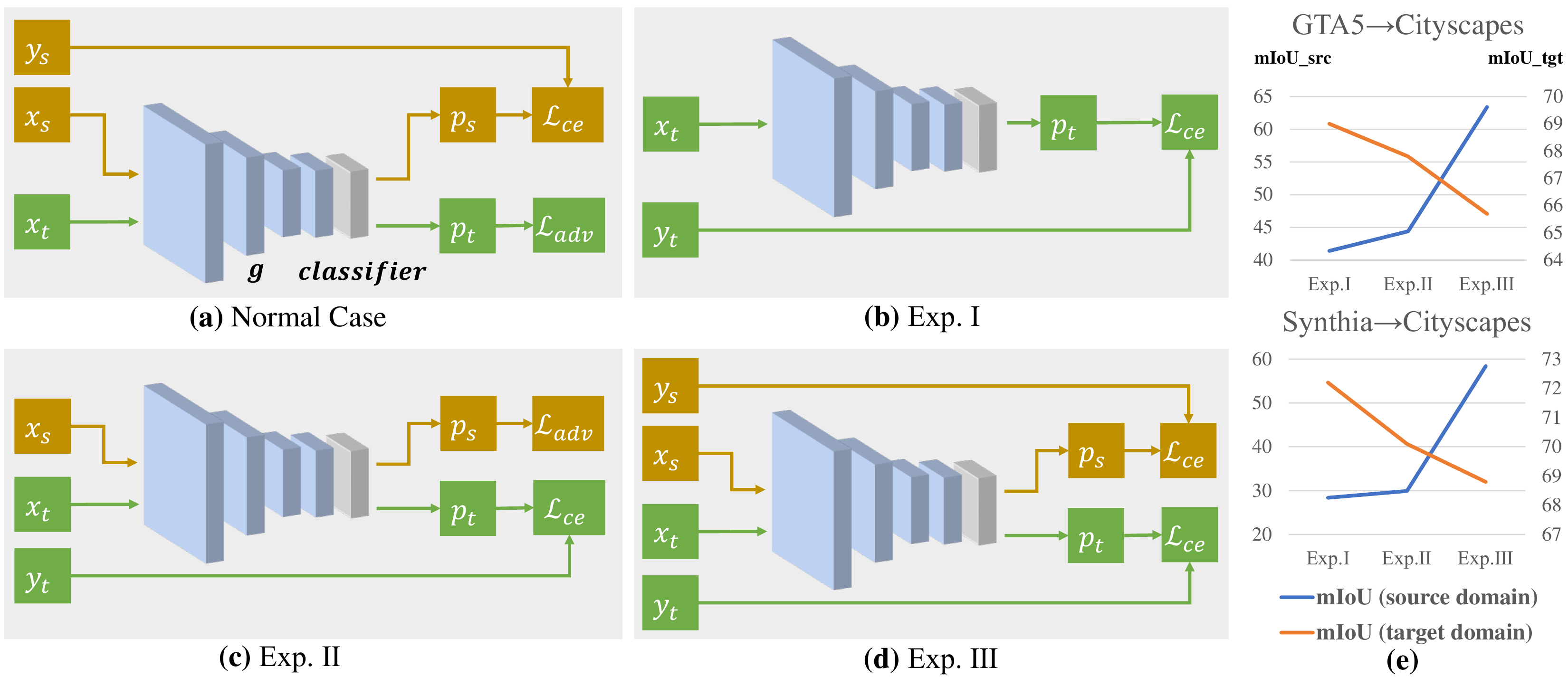}
\end{center}
\caption{(a) A representative domain-invariant method, AdaptSegNet~\cite{adaptnet}. The normal case: the ground-truth labels for the target domain are not available. The brown and green lines represent the source and target domain branches, respectively. (b) Exp.~\uppercase\expandafter{\romannumeral1}: training by target domain images and their ground-truth labels with CE loss. (c) Exp.~\uppercase\expandafter{\romannumeral2}: training by source domain images with the adversarial loss, as well as the target domain images and labels with CE loss. The discriminator is not shown in the figure. (d) Exp.~\uppercase\expandafter{\romannumeral3}: training by both source and target domains images and labels with two CE losses. Note that unlike the normal case in (a), we use target domain ground-truth labels in the toy experiments only to support our idea rather than give a solution. (e) Evaluation results of two benchmarks on both source (blue line) and target (orange line) domain validation sets. Best viewed in color}
\label{fig:toy_exp}
\end{figure}

\subsection{Preliminary}
\label{sec:preliminary}

\subsubsection{Problem definition.}Formally, we have the source domain images $\mathcal X_s$ along with ground-truth labels $\mathcal Y_s$, and the unlabeled target domain images $\mathcal X_t$. Our goal is to train a segmentation model $\mathcal G$ that performs well on the target domain.

A representative domain-invariant solution~\cite{adaptnet} is shown in Fig.~\ref{fig:toy_exp} (a). The source and target domain images $(x_s, x_t)$ pass forward the segmentation network $\mathcal G$, which is typically composed of an encoder $\boldsymbol g$ and a classifier $\mathcal C$, to obtain the predictions $(p_s, p_t)$, respectively. It is written as

\begin{footnotesize}
\begin{equation}
    p_s = \mathcal C(\boldsymbol g(x_s)) ,\quad p_t = \mathcal C(\boldsymbol g(x_t)).
\end{equation}
\end{footnotesize}

For the source domain prediction $p_s$, the cross-entropy loss is employed with its ground-truth label $y_s$ as

\begin{equation}
\footnotesize
\label{eq:ce_loss}
    \mathcal L_{ce} = -\frac{1}{N} \sum_{i=1}^{N} \sum_{c=1}^{C} \textbf{1}\{y_{s,i} = c\} \log p_{s,i,c} ,
\end{equation}
where $N$ is the number of spatial locations in the source prediction map $p_s$, $C$ is the number of classes, $y_{s,i}$ represents the class label at the $i$-th location, and $p_{s,i,c}$ represents the source prediction score of the $c$-th class at the $i$-th location.

As for the target domain prediction $p_t$, a discriminator $\mathcal D$ is used to align the distributions of the source and target predictions. The adversarial loss $\mathcal L_{adv}$ is defined as
\begin{equation}
\footnotesize
\label{eq:adv_loss}
    \mathcal L_{adv} = \frac{1}{N_d} \sum_{i=1}^{N_d} (\mathcal D(p_t)_i - 0)^2,
\end{equation}
where $N_d$ is the number of spatial locations in the discriminator output, $0$ is the label of the source domain, and we follow LSGAN~\cite{lsgan} to use the MSE Loss.

The final loss $\mathcal L_{seg}$ for the segmentation network is defined as
\begin{equation}
\footnotesize
    \mathcal L_{seg} = \mathcal L_{ce} + \lambda_{adv} \mathcal L_{adv},
\end{equation}
where $\lambda_{adv}$ controls the weight for $\mathcal L_{adv}$.
To train the discriminator, the discriminator loss $\mathcal L_d$ is defined as
\begin{equation}
\footnotesize
    \mathcal L_d = \frac{1}{N_d} \sum_{i=1}^{N_d} (\mathcal D(p_s)_i - 0)^2 + \frac{1}{N_d} \sum_{i=1}^{N_d} (\mathcal D(p_t)_i - 1)^2,
\end{equation}
where the labels of source and target domain are $0$ and $1$, respectively. Training alternates between updating the segmentation network with $\mathcal L_{seg}$ and the discriminator with $\mathcal L_d$. 

\subsection{Motivation}
\label{sec:motivation}

The method above aligns the distributions of source and target domain features for domain-invariant learning. However, as shown in Fig.~\ref{fig:toy_exp}(a), since the learning objective involves $\mathcal L_{ce}$ during training, the trained network has to fit the source domain data very well. The source domain overfitting issue potentially impairs the segmentation performance on the target domain.


We conduct three experiments to verify this fact, and show them in Fig.~\ref{fig:toy_exp}(b)-(d). Unlike the normal case (Fig.~\ref{fig:toy_exp}(a)), we use the target domain ground-truth labels in the toy experiments only to support our idea rather than give a solution. 
As shown in Fig.~\ref{fig:toy_exp}(e), from Exp.~\uppercase\expandafter{\romannumeral1} to \uppercase\expandafter{\romannumeral2}, we apply an extra adversarial loss, so the model performs slightly better on the source domain data. Further, from Exp.~\uppercase\expandafter{\romannumeral2} to \uppercase\expandafter{\romannumeral3}, we apply a stronger CE loss on the source domain, so it performs very well on the source domain. However, the results in Fig.~\ref{fig:toy_exp}(e) reveal the fact that the better the model fits on the source domain data, the worse it performs on the target domain. This exactly supports our idea, i.e., overfitting the source domain data actually impairs the final performance on the target domain.

Motivated by the observations, we propose a new framework to decouple the feature distribution alignment from the segmentation task. It alleviates the issue of source domain overfitting, and enables the final model to focus more on target-domain semantic segmentation.

\subsection{DecoupleNet}
\label{sec:arch}

\begin{figure}[t]
\begin{center}
\includegraphics[width=0.99\linewidth]{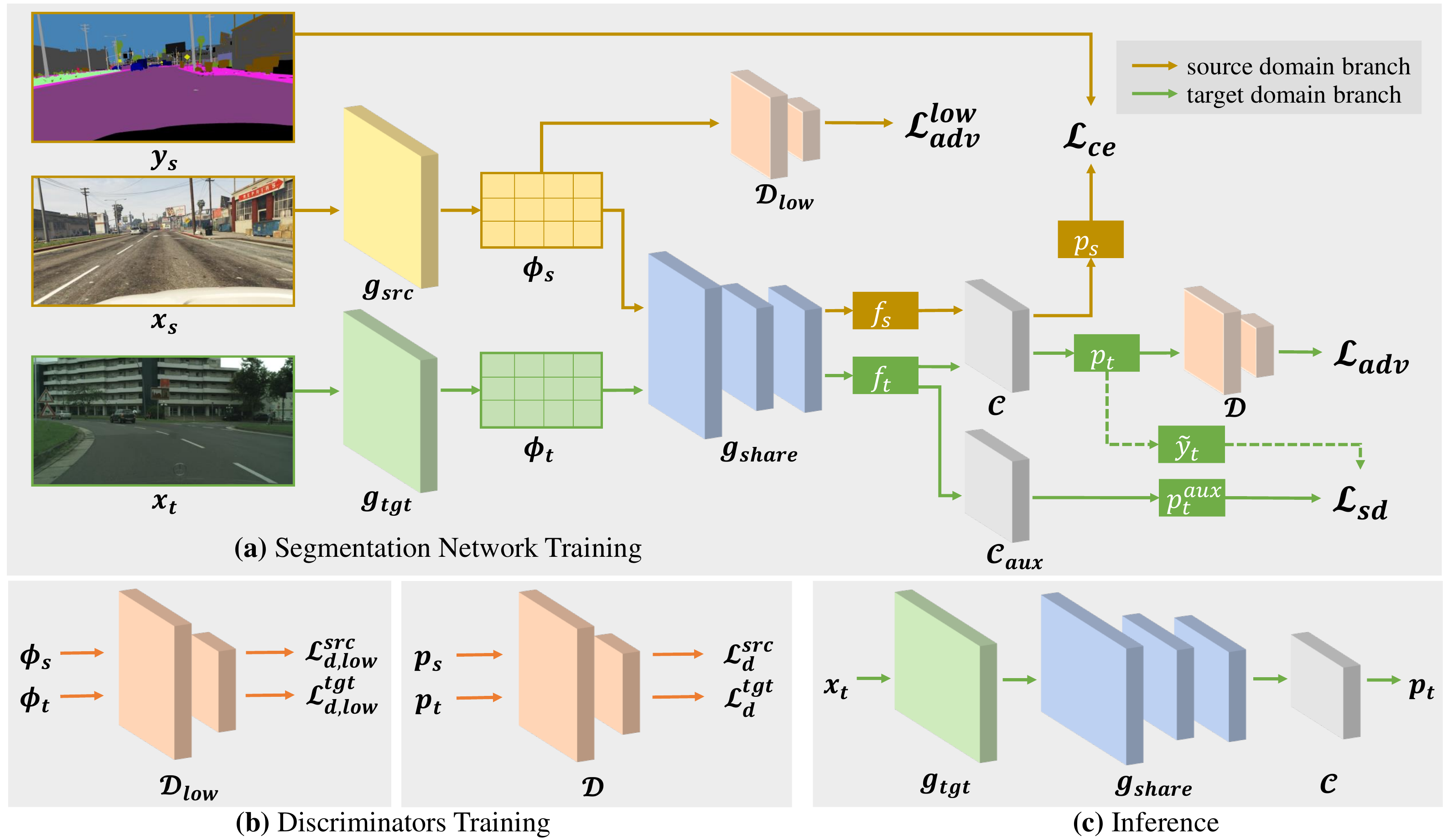}
\end{center}
\caption{Framework of DecoupleNet. (a) Segmentation network training. The brown line represents the source branch, while the green line denotes the target branch. Note that the dashed line means stopping gradients. (b) Discriminators training. (c) Inference pipeline. Best viewed in color}
\label{fig:overview}
\end{figure}


The framework of DecoupleNet is shown in Fig.~\ref{fig:overview}. We first split the feature encoder $\boldsymbol g$ into two parts, i.e., $\boldsymbol g_{tgt}$ and $\boldsymbol g_{share}$. Besides, we maintain another module $\boldsymbol g_{src}$, which shares the same architecture with $\boldsymbol g_{tgt}$. The source and target domain images $(x_s, x_t)$ are fed into the source blocks $\boldsymbol g_{src}$ and target blocks $\boldsymbol g_{tgt}$ to yield the shallow features $(\phi_s,\phi_t)$, respectively. They further pass through the shared blocks $\boldsymbol g_{share}$ to get the features $(f_s,f_t)$. Afterwards, they are passed into the classifier $\mathcal C$ to obtain the predictions $(p_s,p_t)$. Initially, we have
\begin{equation}
\footnotesize
    \phi_s = \boldsymbol g_{src}(x_s),\quad f_s = \boldsymbol g_{shared}(\phi_s), \quad p_s = \mathcal C(f_s),
\end{equation}

\begin{equation}
\footnotesize
    \phi_t = \boldsymbol g_{tgt}(x_t), \quad f_t = \boldsymbol g_{shared}(\phi_t), \quad p_t = \mathcal C(f_t).
\end{equation}

Then, we adopt cross-entropy loss $\mathcal L_{ce}$ for the labeled source domain data as
\begin{equation}
\footnotesize
    \mathcal L_{ce} = -\frac{1}{N} \sum_{i=1}^{N} \sum_{c=1}^{C} \textbf{1}\{y_{s,i} = c\} \log p_{s,i,c}.
\end{equation}

 

Besides, we require the distribution of the source-domain shallow features $\phi_{s}$ to align towards that of the target domain, i.e., $\phi_{t}$, since our goal is to let the source blocks $\boldsymbol g_{src}$ bear the responsibility of feature distribution alignment. Specifically, adversarial learning is adopted for the shallow feature alignment with an additional discriminator $\mathcal D_{low}$ and an adversarial loss $\mathcal L_{adv}^{low}$ as 
\begin{equation}
\footnotesize
\label{eq:adv_low_loss}
    \mathcal L_{adv}^{low} = \frac{1}{N_d^{low}} \sum_{i=1}^{N_d^{low}} (\mathcal D_{low}(\phi_s)_i - 1)^2,
\end{equation}
where $N_d^{low}$ denotes the number of locations in the discriminator output, and $1$ is the label of the target domain.

The design of DecoupleNet is with the following consideration. Basically, the source domain images differ from the target ones mainly in low-level information, such as illumination and texture. Also, it is known that the shallow layers in a network often do well in capturing the low-level information. 
With these two facts, it is natural to let the source blocks $ g_{src}$ align the source-domain shallow features towards the target ones.

Practically, the shallow feature distribution alignment by $\mathcal L_{adv}^{low}$ may be imperfect, and the shallow features for the source and target domains may still be slightly mismatched. Therefore, we also use the adversarial loss $\mathcal L_{adv}$ in the output space, as defined in Eq. \eqref{eq:adv_loss}. In this way, we have the final loss $\mathcal L_{seg}$ as follows for training the segmentation network.
\begin{equation}
\footnotesize
    \mathcal L_{seg} = \mathcal L_{ce} + \lambda_{adv}^{low} \mathcal L_{adv}^{low} + \lambda_{adv} \mathcal L_{adv},
\end{equation}
where $\lambda_{adv}^{low}$ and $\lambda_{adv}$ control the contributions of the corresponding loss. It is notable that the incorporation of $\mathcal{L}_{adv}$ only brings minor improvement (+0.3\% mIoU), as shown in Exp. 5 and 6 of Table~\ref{table:ablation}. This shows the feature alignment is mainly attributed to $\mathcal L_{adv}^{low}$. The $\mathcal L_{adv}$ only serves as a complement.

To train the discriminators, as shown in Fig.~\ref{fig:overview}(b), we follow the previous work~\cite{adaptnet} to yield the discriminator loss as
\begin{equation}
\footnotesize
    \mathcal L_d^{low} = \frac{1}{N_d^{low}} \sum_{i=1}^{N_d^{low}} (\mathcal D_{low}(\phi_s)_i - 0)^2 + \frac{1}{N_d^{low}} \sum_{i=1}^{N_d^{low}} (\mathcal D_{low}(\phi_t)_i - 1)^2,
\end{equation}
\begin{equation}
    \mathcal L_d = \frac{1}{N_d} \sum_{i=1}^{N_d} (\mathcal D(p_s)_i - 0)^2 + \frac{1}{N_d} \sum_{i=1}^{N_d} (\mathcal D(p_t)_i - 1)^2.
\end{equation}

During inference, as shown in Fig.~\ref{fig:overview}(c), we adopt $\mathcal F =\mathcal{C} \circ g_{share} \circ g_{tgt}$ as the final model. All other modules are simply discarded. 
Note that we do not introduce extra parameters during inference.

\subsubsection{Advantage of DecouleNet.}
(1) The source blocks $\boldsymbol g_{src}$ now bear the responsibility of feature distribution alignment. Being less distracted by feature alignment, the final model (i.e., $\boldsymbol g_{share} \circ \boldsymbol g_{tgt}$) focuses more on the segmentation task. (2) Though the source domain branch $\boldsymbol g_{share} \circ \boldsymbol g_{src}$ needs to directly fit the source domain data with $\mathcal L_{ce}$, the final model $\boldsymbol g=\boldsymbol g_{share} \circ \boldsymbol g_{tgt}$ is never required to perform well on the source domain during training. This alleviates the source domain overfitting problem and facilitates performance enhancement on the target domain, as shown in Fig.~\ref{fig:intro}(c).

\subsection{Self-Discrimination}
\label{sec:sd}

Despite the effectiveness of DecoupleNet, the target blocks $\boldsymbol g_{tgt}$ is updated only according to $\mathcal L_{adv}$, which may not be strong enough to learn optimal parameters for $\boldsymbol g^{tgt}$. Also, without a proper learning objective for the target domain, the features $f_t$ may not be discriminative enough. To this end, we propose Self-Discrimination (SD) to provide more supervision on the target domain branch.

As shown in Fig.~\ref{fig:overview}(a), we introduce an auxiliary classifier $\mathcal C_{aux}$, which shares the same architecture with the main classifier $\mathcal C$. As the target domain feature $f_t$ passes the classifier $\mathcal C$ to yield $p_t$, we also forward $f_t$ into the auxiliary classifier $\mathcal C^{aux}$ to get the auxiliary prediction $p_t^{aux}$. Meanwhile, we calculate the pseudo label $\tilde{y}_t$ according to the main prediction $p_t$. Similar to \cite{zou2018unsupervised}, we adopt class-wise thresholds $\boldsymbol \tau$ to ignore uncertain pixels in the pseudo labels and maintain class balancing as well. Finally, we yield the self-discrimination loss $\mathcal L_{sd}$ as
\begin{equation}
\footnotesize
    p_{t}^{aux} = \mathcal C_{aux}(f_t) ,\quad \hat{y}_{t,i} = \mathop{\text{argmax}}_{c=1}^C  p_{t,i,c}, \quad 
    \tilde{y}_{t,i}=
  \begin{cases}
  \hat{y}_{t,i} & p_{t,i,c=\hat{y}_{t,i}} \ge \boldsymbol \tau_{c=\hat{y}_{t,i}} \\
  -1 (ignored) & otherwise
  \end{cases}, \notag
\end{equation}


\begin{equation}
\footnotesize
    \mathcal L_{sd} = -\frac{1}{N_{t}} \sum_{i=1}^{N_{t}} \sum_{c=1}^{C} \textbf{1}\{\tilde{y}_{t,i} = c\} \log p_{t,i,c}^{aux},
\end{equation}
where $C$ is the number of classes, $N_{t}$ is the number of spatial locations in the auxiliary prediction map $p_t^{aux}$, $\boldsymbol \tau_c$ is the threshold for the $c$-th class, $p_{t,i,c}$ and $p_{t,i,c}^{aux}$ are the main and auxiliary prediction scores of the $c$-th class for the feature at the $i$-th location, respectively, and $\tilde{y}_{t,i}$ is the pseudo label for the $i$-th location in the prediction map $p_t$. 

It is notable that the class-wise thresholds $\boldsymbol \tau$ are initialized to zero at the start of training. Then, it is updated with the current predictions $p_t$ at each iteration. The implementation details are given in the supplementary material.

Basically, $\mathcal L_{sd}$ is a cross-entropy loss applied to $p_{t,i,c}^{aux}$. It has a nice property that can adaptively scale the gradients with the current prediction error. Hence, it is capable of yielding more discriminative target features $f_t$. To verify the effectiveness, we compare the t-SNE visualizations with and without SD in the supplementary material. 
During inference, we only use the main classifier, and the auxiliary classifier is simply discarded.


\begin{wrapfigure}{r}{0.46\textwidth}
\centering
\includegraphics[width=0.95\linewidth]{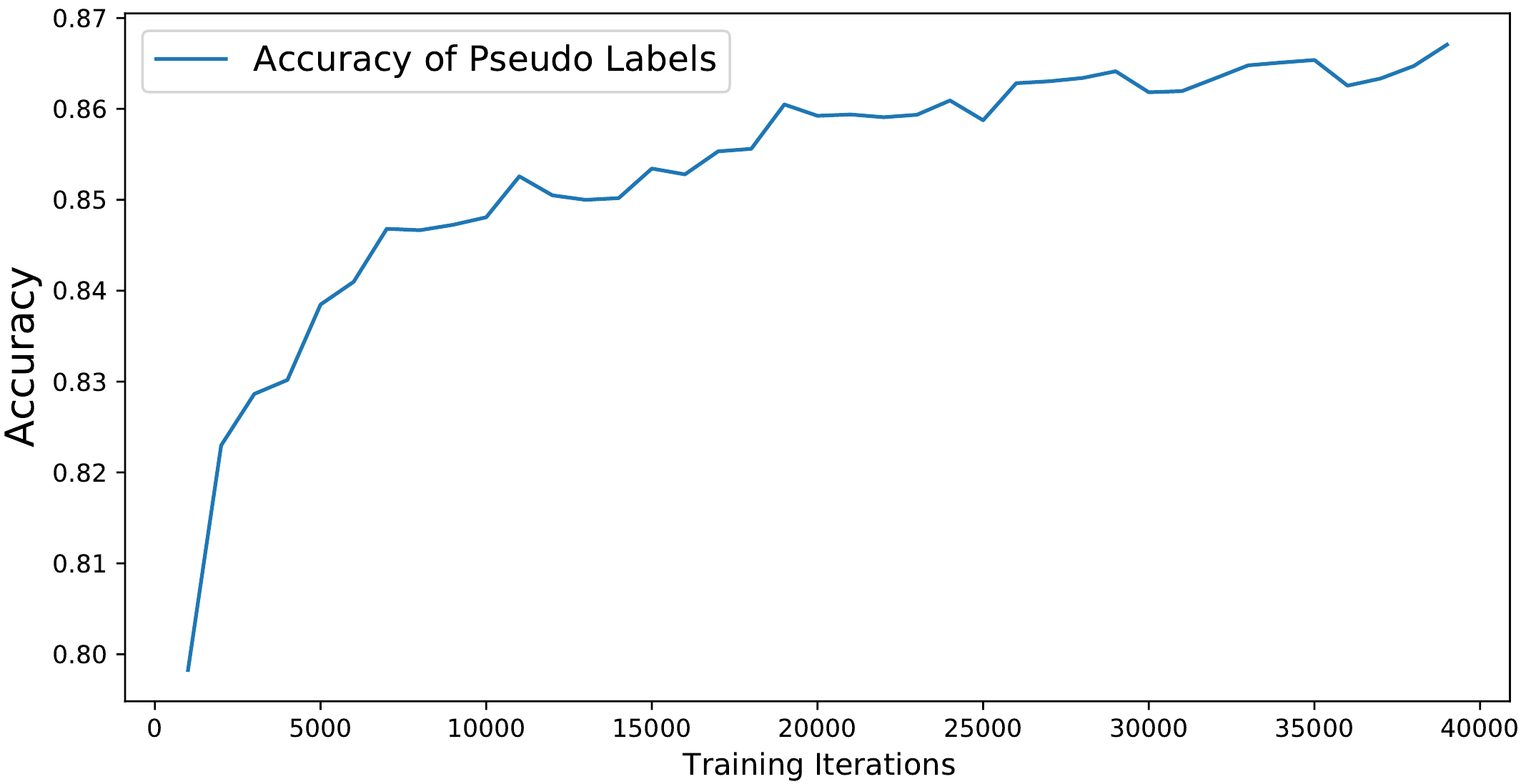}
\caption{Accuracy of pseudo labels during training}
\label{fig:acc}
\end{wrapfigure}

Remarkably, the accuracy of the pseudo labels is more than 80\%, and continues to increase during training, as shown in Fig.~\ref{fig:acc}. Therefore, although there might be wrong supervision from pseudo labels, the benefits brought by SD still outweigh the risks.

Finally, we incorporate $\mathcal L_{sd}$ into the final segmentation loss $\mathcal L_{seg}$ as
\begin{equation}
\footnotesize
    \mathcal L_{seg} = \mathcal L_{ce} + \lambda_{adv}^{low} \mathcal L_{adv}^{low} + \lambda_{adv} \mathcal L_{adv} + \lambda_{sd} \mathcal L_{sd}.\notag
\end{equation}
\subsubsection{The auxiliary classifier.}The auxiliary classifier plays an important role in SD. If we directly apply the self-discrimination loss $\mathcal L_{sd}$ on the main prediction $p_t$ without the auxiliary classifier, the noisy pseudo labels may corrupt the normal training of the main classifier with $\mathcal L_{ce}$ and cause large performance degradation, as shown in Exp. 1 and 2 of Table~\ref{table:ablation_sd}. In contrast, introducing an auxiliary classifier avoids the side effect on the main classifier. 

\subsection{Online Enhanced Self-Training}
\label{sec:oest}

\begin{figure}[t]
\begin{center}
\includegraphics[width=0.9\linewidth]{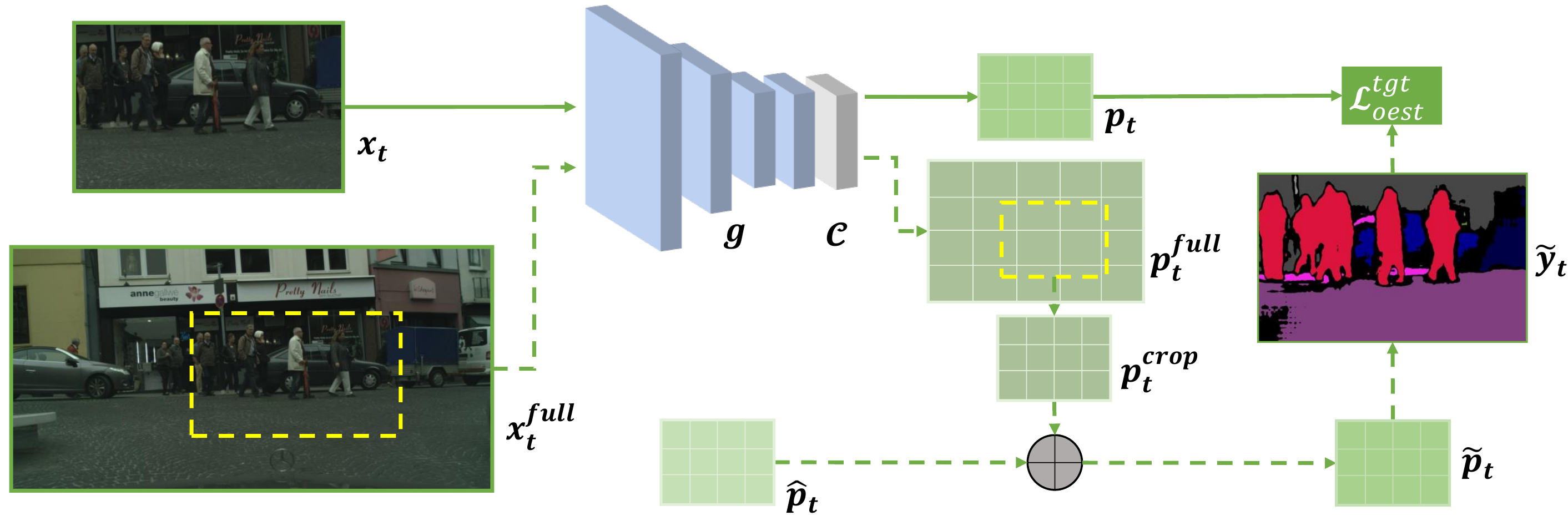}
\end{center}
\caption{Framework of Online Enhanced Self-Training. Dashed line: stopping gradients} 
\label{fig:oest}
\end{figure}

To further boost performance, we extend DecoupleNet from a single stage to a multi-stage self-training paradigm. Most existing self-training-based methods~\cite{zou2018unsupervised,shin2020two,mei2020instance,zhang2021prototypical,yang2020label,kim2020learning,fada} generate pseudo labels in the re-labeling phase and directly use them to provide supervision without further update in the re-training phase. Generally, the predictions get more and more accurate in the re-training process, so fixing the generated pseudo labels may lead to inferior performance. ProDA~\cite{zhang2021prototypical} uses prototypes to denoise the pseudo labels. But it requires to maintain an extra momentum encoder and needs to update the prototypes at each iteration. In contrast, we propose a simple yet effective method, i.e., Online Enhanced Self-Training (OEST), to contextually enhance the pseudo labels via a simple average operation at each iteration.

The framework of OEST is given in Fig.~\ref{fig:oest}. After the first-stage training explained in Sections~\ref{sec:arch} and~\ref{sec:sd}, we generate the pseudo soft labels $\hat{p}_{t} \in [0, 1]^{H \times W \times C}$ by making predictions on each target domain training image $x_{t}$ using the trained model. Then, in the re-training process, we pass the target domain image crops $x_t$ with strong data augmentation (e.g., color jitter) into the segmentation network $\mathcal G=\mathcal C \circ \boldsymbol g$ to yield their predictions $p_t$. In addition, we forward their corresponding full images $x_t^{full}$ with weak data augmentation (e.g., random horizontal flip) to obtain the full predictions $p_t^{full}$ as
\begin{equation}
\footnotesize
    p_t = \mathop{softmax}(\mathcal G(x_t)), \quad p_t^{full} = \mathop{softmax}(\mathcal G(x_t^{full})).
\end{equation}

Afterwards, we crop $p_t^{crop}$ from $p_t^{full}$ in the way as cropping $x_t$ from $x_t^{full}$, and enhance the original pseudo soft labels $\hat{p}_t$ with $p_t^{crop}$ via a simple average operation to yield $\tilde{p}_t$. It follows by ignoring uncertain pixels with class-wise thresholds $\boldsymbol \tau^{st}$ as in \cite{zou2018unsupervised} to obtain the updated pseudo labels $\tilde{y}_t$ as
\begin{equation}
\footnotesize
    \tilde{p}_t = \frac{1}{2}  (\hat{p}_t + p_t^{crop}),\quad \hat{y}_{t,i} = \mathop{\text{argmax}}_{c=1}^C  \tilde{p}_{t,i,c}, \quad 
    \tilde{y}_{t,i}=
  \begin{cases}
  \hat{y}_{t,i} & \tilde{p}_{t,i,c=\hat{y}_{t,i}} \ge \boldsymbol \tau^{st}_{c=\hat{y}_{t,i}} \\
  -1 (ignored) & otherwise
  \end{cases}. \notag
\end{equation}

Since $p_t^{crop}$ is aware of the contexts in the full image, the quality of the original pseudo labels can be contextually enhanced via the simple fusion. Finally, we yield the self-training loss $\mathcal L_{oest}^{tgt}$ on $p_t$ with $\tilde{y}_t$, and add it to the source domain CE loss $\mathcal L_{ce}^{src}$ to obtain the final loss $\mathcal L$ as
\begin{equation}
\footnotesize
\mathcal L_{oest}^{tgt} = -\frac{1}{N_{t}} \sum_{i=1}^{N_{t}} \sum_{c=1}^{C} \textbf{1}\{\tilde{y}_{t,i} = c\} \log p_{t,i,c}, \quad \mathcal L = \mathcal L_{ce}^{src} + \mathcal L_{oest}^{tgt}.
\end{equation}

\section{Experiment}


\subsection{Implementation Details}
\label{sec:setup}

\subsubsection{Experimental setting.}
Following previous work~\cite{adaptnet,fada,clan,advent,luo2019significance,adapt_patch,ssf}, we use the ResNet-101~\cite{resnet} and DeepLabv2~\cite{deeplabv2} as our base model. 
To split the feature encoder, we take \{layer$0$, layer$1$\} as the target blocks $\boldsymbol{g}_{tgt}$ and the rest as the shared blocks $\boldsymbol{g}_{share}$ for GTA5 dataset, while \{layer$0$, layer$1$, layer$2$\} as $\boldsymbol{g}_{tgt}$ and the rest as $\boldsymbol{g}_{share}$ for Synthia dataset. Note that layer$0$ refers to \{conv1, bn1, relu, maxpool\}. More details are given in the supplementary material. 

\subsubsection{Datasets.}
Following most previous work, evaluation is performed on GTA5 $\rightarrow$ Cityscapes, Synthia $\rightarrow$ Cityscapes and Cityscapes $\rightarrow$ Cross-City. The details of the datasets~\cite{gta5,synthia,cityscapes,chen2017no} are given in the supplementary material.
\begin{table}[t]
\tabcolsep=0.083cm
\caption{Results on GTA5$\rightarrow$Cityscapes with ResNet101 and DeepLabv2. ST: self-training}
\label{table:comparison_gta}
\begin{center}
\resizebox{\textwidth}{!}{
\begin{tabular}{l | c | c c c c c c c c c c c c c c c c c c c | c }
\hline
\toprule Method & ST & \rotatebox{70}{road} & \rotatebox{70}{sw.} & \rotatebox{70}{build} & \rotatebox{70}{wall} & \rotatebox{70}{fence} & \rotatebox{70}{pole} & \rotatebox{70}{light} & \rotatebox{70}{sign} & \rotatebox{70}{veg.} & \rotatebox{70}{terrain} & \rotatebox{70}{sky} & \rotatebox{70}{person} & \rotatebox{70}{rider} & \rotatebox{70}{car} & \rotatebox{70}{truck} & \rotatebox{70}{bus} & \rotatebox{70}{train} & \rotatebox{70}{moto.} & \rotatebox{70}{bicycle} & mIoU \\
\specialrule{0em}{2pt}{0pt}
\hline
\specialrule{0em}{2pt}{0pt}

 SourceOnly &  & 27.0 & 20.6 & 53.9 & 20.8 & 19.4 & 35.3 & 40.7 & 23.0 & 84.6 & 30.1 & 73.5 & 63.9 & 31.4 & 65.7 & 10.5 & 26.3 & 2.1 & 34.1 & 21.8 & 36.0\\

 AdaptSeg~\cite{adaptnet} &  & 86.5 & 36.0 & 79.9 & 23.4 & 23.3 & 23.9 & 35.2 & 14.8 & 83.4 & 33.3 & 75.6 & 58.5 & 27.6 & 73.7 & 32.5 & 35.4 & 3.9 & 30.1 & 28.1 & 42.4 \\

 AdaptSeg\textit{(LS)} &  & \textbf{91.4} & \textbf{48.4} & 81.2 & 27.4 & 21.2 & 31.2 & 35.3 & 16.1 & 84.1 & 32.5 & 78.2 & 57.7 & 28.2 & 85.9 & 33.8 & 43.5 & 0.2 & 23.9 & 16.9 & 44.1\\

CLAN~\cite{clan} &  & 87.0 & 27.1 & 79.6 & 27.3 & 23.3 & 28.3 & 35.5 & 24.2 & 83.6 & 27.4 & 74.2 & 58.6 & 28.0 & 76.2 & 33.1 & 36.7 & 6.7 & 31.9 & 31.4 & 43.2 \\

 AdvEnt~\cite{advent} &  & 89.4 & 33.1 & 81.0 & 26.6 & 26.8 & 27.2 & 33.5 & 24.7 & 83.9 & \textbf{36.7} & 78.8 & 58.7 & 30.5 & 84.8 & \textbf{38.5} & \textbf{44.5} & 1.7 & 31.6 & 32.4 & 45.5 \\
 
 FADA~\cite{fada} & & 88.5 & 39.7 & \textbf{83.6} & \textbf{37.9} & 24.7 & 27.5 & 34.1 & 21.3 & 83.3 & 32.9 & \textbf{83.4} & 58.0 & 33.5 & 84.7 & 37.9 & 39.8 & 25.2 & 30.8 & 27.6 & 47.1 \\
 
  
  
  \rowcolor{Gray}Ours &  & 87.5 & 37.6 & 83.2 & 31.6 & \textbf{28.3} & \textbf{38.6} & \textbf{44.3} & \textbf{24.9} & \textbf{85.1} & 31.0 & 76.0 & \textbf{68.1} & \textbf{36.9} & \textbf{86.4} & 28.4 & 39.0 & \textbf{25.5} & \textbf{42.8} & \textbf{36.1} & \textbf{49.0} \\
  
\specialrule{0em}{2pt}{0pt}
\hline
\specialrule{0em}{2pt}{0pt}

  CBST~\cite{zou2018unsupervised} & \Checkmark & 91.8 & 53.5 & 80.5 & 32.7 & 21.0 & 34.0 & 28.9 & 20.4 & 83.9 & 34.2 & 80.9 & 53.1 & 24.0 & 82.7 & 30.3 & 35.9 & 16.0 & 25.9 & 42.8 & 45.9 \\
  
 AdaptPatch~\cite{adapt_patch} & \Checkmark & 92.3 & 51.9 & 82.1 & 29.2 & 25.1 & 24.5 & 33.8 & 33.0 & 82.4 & 32.8 & 82.2 & 58.6 & 27.2 & 84.3 & 33.4 & 46.3 & 2.2 & 29.5 & 32.3 & 46.5 \\
 

 Label-Driven~\cite{yang2020label} & \Checkmark & 90.8 & 41.4 & 84.7 & 35.1 & 27.5 & 31.2 & 38.0 & 32.8 & 85.6 & 42.1 & 84.9 & 59.6 & 34.4 & 85.0 & \textbf{42.8} & 52.7 & 3.4 & 30.9 & 38.1 & 49.5\\

 FADA~\cite{fada} & \Checkmark & 91.0 & 50.6 & 86.0 & \textbf{43.4} & 29.8 & 36.8 & 43.4 & 25.0 & 86.8 & 38.3 & 87.4 & 64.0 & 38.0 & 85.2 & 31.6 & 46.1 & 6.5 & 25.4 & 37.1 & 50.1 \\
 
 Kim et al.~\cite{kim2020learning} & \Checkmark & 92.9 & 55.0 & 85.3 & 34.2 & 31.1 & 34.9 & 40.7 & 34.0 & 85.2 & 40.1 & 87.1 & 61.0 & 31.1 & 82.5 & 32.3 & 42.9 & 0.3 & 36.4 & 46.1 & 50.2 \\

 FDA-MBT~\cite{yang2020fda} & \Checkmark & 92.5 & 53.3 & 82.4 & 26.5 & 27.6 & 36.4 & 40.6 & 38.9 & 82.3 & 39.8 & 78.0 & 62.6 & 34.4 & 84.9 & 34.1 & 53.1 & 16.9 & 27.7 & 46.4 & 50.5 \\
 
 TPLD~\cite{shin2020two} & \Checkmark & \textbf{94.2} & \textbf{60.5} & 82.8 & 36.6 & 16.6 & 39.3 & 29.0 & 25.5 & 85.6 & \textbf{44.9} & 84.4 & 60.6 & 27.4 & 84.1 & 37.0 & 47.0 & 31.2 & 36.1 & 50.3 & 51.2 \\
 
 IAST~\cite{mei2020instance} & \Checkmark & 94.1 & 58.8 & 85.4 & 39.7 & 29.2 & 25.1 & 43.1 & 34.2 & 84.8 & 34.6 & 88.7 & 62.7 & 30.3 & 87.6 & 42.3 & 50.3 & 24.7 & 35.2 & 40.2 & 52.2 \\
 
 MetaCorrection~\cite{guo2021metacorrection} & \Checkmark & 92.8 & 58.1 & 86.2 & 39.7 & 33.1 & 36.3 & 42.0 & 38.6 & 85.5 & 37.8 & 87.6 & 62.8 & 31.7 & 84.8 & 35.7 & 50.3 & 2.0 & 36.8 & 48.0 & 52.1 \\
 
 DPL~\cite{cheng2021dual} & \Checkmark & 92.8 & 54.4 & 86.2 & 41.6 & 32.7 & 36.4 & 49.0 & 34.0 & 85.8 & 41.3 & 86.0 & 63.2 & 34.2 & 87.2 & 39.3 & 44.5 & 18.7 & 42.6 & 43.1 & 53.3 \\
 
 ProDA~\cite{zhang2021prototypical} & \Checkmark & 91.5 & 52.3 & 82.9 & 41.8 & 35.7 & 40.3 & 44.3 & \textbf{43.2} & 87.1 & 43.4 & 79.6 & 66.6 & 31.6 & 86.9 & 40.1 & 53.0 & 0.0 & 45.7 & 53.2 & 53.6\\
  
 \rowcolor{Gray}Ours+ST & \Checkmark & 88.5 & 47.8 & \textbf{87.4} & 38.3 & \textbf{36.9} & \textbf{44.9} & \textbf{53.8} & 39.6 & \textbf{88.0} & 38.7 & \textbf{88.8} & \textbf{70.4} & \textbf{39.4} & \textbf{87.8} & 31.4 & \textbf{55.0} & \textbf{37.4} & \textbf{47.1} & \textbf{55.9} & \textbf{56.7}\\
 
 
 
\specialrule{0em}{2pt}{0pt}
\hline
\specialrule{0em}{2pt}{0pt}

 ProDA \textit{(w/ SimCLR)} & \Checkmark & \textbf{87.8} & \textbf{56.0} & 79.7 & \textbf{46.3} & \textbf{44.8} & 45.6 & 53.5 & \textbf{53.5} & 88.6 & \textbf{45.2} & 82.1 & 70.7 & 39.2 & \textbf{88.8} & \textbf{45.5} & \textbf{59.4} & 1.0 & 48.9 & 56.4 & 57.5 \\
  
 \rowcolor{Gray}Ours \textit{(w/ SimCLR)} & \Checkmark & 87.6 & 49.3 & \textbf{87.2} & 42.5 & 41.6 & \textbf{46.6} & \textbf{57.4} & 44.0 & \textbf{89.0} & 43.9 & \textbf{90.6} & \textbf{73.0} & \textbf{43.8} & 88.1 & 32.9 & 53.7 & \textbf{44.3} & \textbf{49.8} & \textbf{57.2} & \textbf{59.1}\\

\bottomrule

\end{tabular}}
\end{center}
\end{table}

\begin{table}[t]
\tabcolsep=0.083cm
\caption{Results on Synthia$\rightarrow$Cityscapes with ResNet101 and DeepLabv2. ST: self-training. mIoU\textsuperscript{+}: mIoU of 13 classes}
\label{table:comparison_synthia}
\begin{center}
\resizebox{\textwidth}{!}{
\begin{tabular}{l | c | c c c c c c c c c c c c c c c c | c c }
\hline
\toprule Method & ST & \rotatebox{70}{road} & \rotatebox{70}{sw.} & \rotatebox{70}{build} & \rotatebox{70}{wall} & \rotatebox{70}{fence} & \rotatebox{70}{pole} & \rotatebox{70}{light} & \rotatebox{70}{sign} & \rotatebox{70}{veg.} & \rotatebox{70}{sky} & \rotatebox{70}{person} & \rotatebox{70}{rider} & \rotatebox{70}{car} & \rotatebox{70}{bus} & \rotatebox{70}{moto.} & \rotatebox{70}{bicycle} & mIoU & mIoU\textsuperscript{+}\\
\specialrule{0em}{2pt}{0pt}
\hline
\specialrule{0em}{2pt}{0pt}
 SourceOnly &  & 59.9 & 24.7 & 57.7 & 6.3 & 0.0 & 32.5 & \textbf{29.7} & 15.0 & 72.8 & 70.8 & 59.2 & 17.7 & 73.0 & 23.0 & 11.6 & 22.6 & 36.0 & 41.4 \\

  AdaptSeg~\cite{adaptnet} &  & 79.2 & 37.2 & 78.8 & 10.5 & 0.3 & 25.1 & 9.9 & 10.5 & 78.2 & 80.5 & 53.5 & 19.6 & 67.0 & 29.5 & 21.6 & 31.3 & 39.5 & 45.9 \\

  AdaptSeg\textit{(LS)} &  & 84.0 & 40.5 & 79.3 & 10.4 & 0.2 & 22.7 & 6.5 & 8.0 & 78.3 & 82.7 & 56.3 & 22.4 & 74.0 & 33.2 & 18.9 & \textbf{34.9} & 40.8 & 47.6 \\

  CLAN~\cite{clan} &  & 81.3 & 37.0 & \textbf{80.1} & - & - & - & 16.1 & 13.7 & 78.2 & 81.5 & 53.4 & 21.2 & 73.0 & 32.9 & 22.6 & 30.7 & - & 47.8 \\ 

  AdvEnt~\cite{advent} &  & \textbf{85.6} & \textbf{42.2} & 79.7 & 8.7 & \textbf{0.4} & 25.9 & 5.4 & 8.1 & 80.4 & \textbf{84.1} & 57.9 & 23.8 & 73.3 & \textbf{36.4} & 14.2 & 33.0 & 41.2 & 48.0 \\
 
 
 \rowcolor{Gray}Ours &  & 77.9 & 38.9 & 74.4 & \textbf{11.9} & 0.2 & \textbf{33.3} & 26.5 & \textbf{17.1} & \textbf{83.6} & 80.0 & \textbf{60.7} & \textbf{26.5} & \textbf{79.9} & 26.4 & \textbf{25.5} & 33.5 & \textbf{43.5} & \textbf{50.1}\\
 
\specialrule{0em}{2pt}{0pt}
\hline
\specialrule{0em}{2pt}{0pt}

  CBST~\cite{zou2018unsupervised} & \Checkmark & 68.0 & 29.9 & 76.3 & 10.8 & 1.4 & 33.9 & 22.8 & 29.5 & 77.6 & 78.3 & 60.6 & 28.3 & 81.6 & 23.5 & 18.8 & 39.8 & 42.6 & 48.9 \\
  
  AdaptPatch~\cite{adapt_patch} & \Checkmark & 82.4 & 38.0 & 78.6 & 8.7 & 0.6 & 26.0 & 3.9 & 11.1 & 75.5 & 84.6 & 53.5 & 21.6 & 71.4 & 32.6 & 19.3 & 31.7 & 40.0 & 46.5 \\
 
  FADA~\cite{fada} & \Checkmark & 84.5 & 40.1 & 83.1 & 4.8 & 0.0 & 34.3 & 20.1 & 27.2 & 84.8 & 84.0 & 53.5 & 22.6 & 85.4 & 43.7 & 26.8 & 27.8 & 45.2 & 52.5\\
  
  Label-Driven~\cite{yang2020label} & \Checkmark & 85.1 & 44.5 & 81.0 & - & - & - & 16.4 & 15.2 & 80.1 & 84.8 & 59.4 & 31.9 & 73.2 & 41.0 & 32.6 & 44.7 & - & 53.1 \\

  Kim et al.~\cite{kim2020learning} & \Checkmark & 79.3 & 35.0 & 73.2 & - & - & - & 19.9 & 24.0 & 61.7 & 82.6 & 61.4 & 31.1 & 83.9 & 40.8 & 38.4 & 51.1 & - & 52.5\\
  
  FDA-MBT~\cite{yang2020fda} & \Checkmark & 79.3 & 35.0 & 73.2 & - & - & - & 19.9 & 24.0 & 61.7 & 82.6 & 61.4 & 31.1 & 83.9 & 40.8 & 38.4 & 51.1 & - & 52.5\\
  
  TPLD~\cite{shin2020two} & \Checkmark & 80.9 & 44.3 & 82.2 & 19.9 & 0.3 & 40.6 & 20.5 & 30.1 & 77.2 & 80.9 & 60.6 & 25.5 & 84.8 & 41.1 & 24.7 & 43.7 & 47.3 & 53.5 \\
  
 MetaCorrection~\cite{guo2021metacorrection} & \Checkmark & \textbf{92.6} & \textbf{52.7} & 81.3 & 8.9 & 2.4 & 28.1 & 13.0 & 7.3 & 83.5 & 85.0 & 60.1 & 19.7 & 84.8 & 37.2 & 21.5 & 43.9 & 45.1 & 52.5 \\
 
  DPL~\cite{cheng2021dual} & \Checkmark & 87.5 & 45.7 & 82.8 & 13.3 & 0.6 & 33.2 & 22.0 & 20.1 & 83.1 & 86.0 & 56.6 & 21.9 & 83.1 & 40.3 & 29.8 & 45.7 & 47.0 & 54.2\\
  
 IAST~\cite{mei2020instance} & \Checkmark & 81.9 & 41.5 & \textbf{83.3} & 17.7 & \textbf{4.6} & 32.3 & 30.9 & 28.8 & 83.4 & 85.0 & 65.5 & 30.8 & 86.5 & 38.2 & 33.1 & 52.7 & 49.8 & 57.0 \\
 
  ProDA~\cite{zhang2021prototypical} & \Checkmark & 87.3 & 44.2 & \textbf{83.3} & 26.6 & 0.3 & 41.8 & 43.8 & \textbf{33.1} & 86.7 & 82.4 & 69.1 & 25.7 & 88.0 & 50.3 & 31.1 & 43.8 & 52.3 & 59.1\\
  
 \rowcolor{Gray}Ours+ST & \Checkmark & 78.7 & 47.4 & 75.7 & \textbf{27.8} & 1.0 & \textbf{43.3} & \textbf{49.1} & 32.6 & \textbf{87.8} & \textbf{87.3} & \textbf{69.3} & \textbf{34.4} & \textbf{88.5} & \textbf{55.0} & \textbf{44.8} & \textbf{58.5} & \textbf{55.1} & \textbf{62.2}\\
 
 
 
\specialrule{0em}{2pt}{0pt}
\hline
\specialrule{0em}{2pt}{0pt}

 ProDA \textit{(w/ SimCLR)} & \Checkmark & \textbf{87.8} & 45.7 & \textbf{84.6} & \textbf{37.1} & 0.6 & 44.0 & \textbf{54.6} & 37.0 & \textbf{88.1} & 84.4 & \textbf{74.2} & 24.3 & 88.2 & 51.1 & 40.5 & 45.6 & 55.5 & 62.0\\
  
 \rowcolor{Gray}Ours \textit{(w/ SimCLR)} & \Checkmark & 77.8 & \textbf{48.6} & 75.6 & 32.0 & \textbf{1.9} & \textbf{44.4} & 52.9 & \textbf{38.5} & 87.8 & \textbf{88.1} & 71.1 & \textbf{34.3} & \textbf{88.7} & \textbf{58.8} & \textbf{50.2} & \textbf{61.4} & \textbf{57.0} & \textbf{64.1}\\

  
 
\bottomrule                         

\end{tabular}
}
\end{center}
\end{table}

\subsection{Results}
\label{sec:comparison}

The comparison with existing state-of-the-art methods is given in Table~\ref{table:comparison_gta}, Table~\ref{table:comparison_synthia}. Clearly, our method outperforms others by a large margin. Previous methods~\cite{adaptnet,adapt_patch,advent,fada} neglect the adverse effect brought by entanglement of feature distribution alignment and the segmentation task. Contrarily, DecoupleNet decouples these two tasks, hence boosting the performance.

What's more, equipped with OEST, our method demonstrates stronger performance. Notably, our method even surpasses ProDA~\cite{zhang2021prototypical} by 3.1 points on GTA5$\rightarrow$Cityscapes and 2.8 points on Synthia$\rightarrow$Cityscapes, achieving a new state of the art. Furthermore, following ProDA to distill the SimCLR~\cite{simclr} initialized student, our method still outperforms ProDA on both benchmarks. On Cityscapes $\rightarrow$ Cross-City, our method also manifests competitive results given in the supplementary material. 

\begin{table}[t]
    \caption{Ablation study for DecoupleNet and SD. Decouple: decoupled network architecture. SD: Self-Discrimination}
    \label{table:ablation}  
    \centering
    \tabcolsep=0.1cm
    {
        \begin{footnotesize}
        \begin{tabular}{ c | l | c  c  c  c  c | c }
            \toprule
            ID & Method & Decouple & $\mathcal L_{ce}$ & $\mathcal L_{adv}$ & $\mathcal L_{adv}^{low}$ & $\mathcal L_{sd}$ & mIoU \\

            \specialrule{0em}{0pt}{1pt}
            \hline
            \specialrule{0em}{0pt}{1pt}
            
            1 & SourceOnly & & \Checkmark & & & & 36.0\\ 
            
            2 & AdaptSegNet & & \Checkmark & \Checkmark & & & 44.1\\ 
            
            3 & AdaptSegNet + SD &  & \Checkmark & \Checkmark &  & \Checkmark & 46.0 \\
            
            4 & DecoupleNet & \Checkmark & \Checkmark & \Checkmark & \Checkmark & & 47.1 \\ 
            
            5 & DecoupleNet + SD (\textit{w/o $\mathcal L_{adv}^{low}$}) & \Checkmark & \Checkmark & \Checkmark &  & \Checkmark & 46.8 \\
            
            6 & DecoupleNet + SD (\textit{w/o $\mathcal L_{adv}$}) & \Checkmark & \Checkmark &  & \Checkmark & \Checkmark & 48.7 \\
            
            7 & DecoupleNet + SD & \Checkmark & \Checkmark & \Checkmark & \Checkmark & \Checkmark & \textbf{49.0} \\
            
            8 & DecoupleNet (\textit{w/o $\mathcal L_{adv}^{low}$}) & \Checkmark & \Checkmark & \Checkmark & & & 44.7 \\
            
            \bottomrule                                   
        \end{tabular}
        \end{footnotesize}
    }    
\end{table}

\subsection{Ablation Study}
\label{sec:ablation}


\subsubsection{DecoupleNet.}By comparing Exp. 2 and 4 in Table~\ref{table:ablation}, we can see DecoupleNet outperforms the domain-invariant method (AdaptSegNet~\cite{adaptnet}) by 3.0\% mIoU, which reveals the effectiveness of DecoupleNet. Note that except the decoupled architecture and $\mathcal L_{adv}^{low}$, Exp. 2 and 4 are kept all the same for fair comparison. 

\begin{table}[t]
    \caption{Ablation study for the decoupled layers, i.e., the architecture of $\boldsymbol{g}_{src}$ or $\boldsymbol{g}_{tgt}$. Note that ResNet has 5 layers in total, and layer0 refers to the stem layer, i.e., \{conv1, bn1, relu, maxpool\}}
    \label{table:ablation_decouple_layers}   
    \centering
    \tabcolsep=0.4cm
    {
        \begin{footnotesize}
        \begin{tabular}{ c | c | c | c }
            \toprule
            Decoupled layers & \{layer0\} & \{layer0,1\} & \{layer0,1,2\} \\

            \specialrule{0em}{0pt}{1pt}
            \hline
            \specialrule{0em}{0pt}{1pt}
            
            mIoU (\%) & 47.7 & \textbf{49.0} & 47.9\\ 
            
            \bottomrule                                   
        \end{tabular}
        \end{footnotesize}
    }    
\end{table}

Notably, we emphasize that $\mathcal L_{adv}$ brings only slight improvement ($+0.3$\% mIoU) by comparing Exp. 6 and 7 in Table~\ref{table:ablation}. On the other hand, $\mathcal L_{adv}^{low}$ brings large performance boost ($+2.2$\% mIoU), through the comparison between Exp. 5 and 7. This shows that the huge performance boost by DecoupleNet mainly comes from the decoupled network architecture and $\mathcal L_{adv}^{low}$, rather than $\mathcal L_{adv}$. $\mathcal L_{adv}$ only serves as a complement for the imperfect alignment by $\mathcal L_{adv}^{low}$.
This demonstrates the effectiveness of DecoupleNet from another perspective.

Besides, we investigate the effect of decoupled layers (i.e., the architecture of $\boldsymbol{g}_{src}$ or $\boldsymbol{g}_{tgt}$) in Table~\ref{table:ablation_decouple_layers}. Making it too shallow leads to insufficient capability for feature alignment, while making it too deep may interfere segmentation.

\begin{table}[H]

    \caption{Ablation study for the alignment direction between $\phi_s$ and $\phi_t$. $\phi_s\rightarrow\phi_t$: applying $\mathcal L_{adv}^{low}$ on $\phi_s$. $\phi_t\rightarrow\phi_s$: applying $\mathcal L_{adv}^{low}$ on $\phi_t$. No alignment: $\lambda_{adv}^{low}=0$}
    \label{table:ablation_align_dir}   
    \centering
    \tabcolsep=0.4cm
    {
        \begin{footnotesize}
        \begin{tabular}{ c | c | c | c }
            \toprule
            Alignment direction & $\phi_s\rightarrow\phi_t$ & $\phi_t\rightarrow\phi_s$ & No alignment \\

            \specialrule{0em}{0pt}{1pt}
            \hline
            \specialrule{0em}{0pt}{1pt}
            
            mIoU (\%) & \textbf{49.0} & 47.3 & 46.8 \\ 
            
            \bottomrule                                   
        \end{tabular}
        \end{footnotesize}
    }    
\end{table}

Also, we highlight the importance of the alignment direction of $\phi_s$ and $\phi_t$ in Table~\ref{table:ablation_align_dir}. $\phi_s\rightarrow\phi_t$ performs the best. We explain that this prevents the segmentation network $\boldsymbol{g} = \boldsymbol{g}_{share} \circ \boldsymbol{g}_{tgt}$ from being distracted by the feature alignment.


\subsubsection{Self-Discrimination.}
\label{sec:ablation_sd}


By comparing Exp. 4 and 7 in Table~\ref{table:ablation}, we observe a performance boost of $1.9$\% mIoU brought by SD, which clearly demonstrates its effectiveness. Also, it is notable that when we directly apply SD on the domain-invaraint method (i.e., AdaptSegNet~\cite{adaptnet}), the performance still continues to improve by a large margin, through the comparison between Exp. 2 and 3 in Table~\ref{table:ablation}. It shows that SD is not limited to DecoupleNet and can serve as a plugin to existing methods by providing an additional supervision.


In addition, 
we show the t-SNE visualizations of the target domain features $f_t$ with and without SD in the supplementary material. It reveals the fact that the model tends to learn more discriminative target domain features with SD. 

\begin{wraptable}{r}{0.43\textwidth}
\caption{Ablation study for class-wise thresholds and the auxiliary classifier. class-balance: class-wise thresholds. aux: auxiliary classifier}
\label{table:ablation_sd}   
\centering
\begin{footnotesize}
    \begin{tabular}{ c | c | c | c  c}
        \toprule
        ID & class-balance & aux & mIoU & $\Delta$ \\

        \specialrule{0em}{0pt}{1pt}
        \hline
        \specialrule{0em}{0pt}{1pt}
        
        1 & \Checkmark & \Checkmark & 49.0 & 0.0 \\
        
        2 & \Checkmark &  & 46.0 & -3.0 \\ 
        
        3 & & \Checkmark & 48.4 & -0.6\\ 
        
        \bottomrule                                   
    \end{tabular}
\end{footnotesize}
\end{wraptable} 


            
            
            
            

Moreover, to show the necessity of the auxiliary classifier, we make comparison in Table~\ref{table:ablation_sd}. For the model w/o auxiliary classifier (Exp. 2), we directly apply $\mathcal L_{sd}$ on the main predictions $p_t$, which leads to large degradation ($-3.0$\% mIoU) compared to Exp. 1. We conjecture that the supervision signal from the noisy pseudo labels may interfere the normal training of the main classifier with source domain ground-truth labels. Further, Exp. 1 and 3 in Table~\ref{table:ablation_sd} show the effectiveness of the class-wise thresholds, since it alleviates the class-imbalance issue on pseudo labels.

\begin{table}[H]
    \caption{Ablation study for OEST. avg \textit{(full)}: average pseudo soft labels and predictions from full images. avg \textit{(crop)}: average pseudo soft labels and predictions from crops. fix: use fixed pseudo soft labels only. pred only: use predictions only}
    \label{table:ablation_oest}   
    \centering
    \tabcolsep=0.3cm
    {
        \begin{footnotesize}
        \begin{tabular}{ c | c | c | c | c }
            \toprule
            Fusion Method & avg \textit{(full)} & avg \textit{(crop)} & fix & pred only\\

            \specialrule{0em}{0pt}{1pt}
            \hline
            \specialrule{0em}{0pt}{1pt}
            
            mIoU (\%) & \textbf{56.7} & 55.2 & 55.6 & 25.8\\
            
            \bottomrule                                   
        \end{tabular}
        \end{footnotesize}
    }    
\end{table}

            

            
            
            

\subsubsection{Online Enhanced Self-Training.}As shown in Table~\ref{table:ablation_oest}, we compare the models with various fusion methods. The comparison between `avg \textit{(full)}' and `avg \textit{(crop)}' show the effectiveness of contextual enhancement via full predictions. Moreover, `fix' is inferior to `avg \textit{(full)}' by 1.1\% mIoU, which shows that online updating pseudo labels with current predictions indeed improves the quality of pseudo labels and brings performance boost. As for `pred only', it totally corrupts the training potentially due to the instability of the online prediction. 

\section{Conclusion}
We have observed two issues of existing domain-invariant learning methods -- (1) \textit{tasks entanglement} and (2) \textit{source domain overfitting}. We propose DecoupleNet to enable the final model to focus more on the segmentation task. Moreover, Self-Discrimination is put forward to learn more discriminative target features. Finally, we design OEST to contextually enhance the pseudo labels.
%
%
\bibliographystyle{splncs04}
\bibliography{egbib}
\end{document}